# On the Place of Text Data in Lifelogs, and Text Analysis via Semantic Facets

Gregory Grefenstette[1], Lawrence Muchemi[2]
[1]INRIA, Saclay, France
[2]University of Nairobi, Kenya & INRIA, Saclay, France

**Abstract**
Current research in lifelog data has not paid enough attention to analysis of cognitive activities in comparison to physical activities. We argue that as we look into the future, wearable devices are going to be cheaper and more prevalent and textual data will play a more significant role. Data captured by lifelogging devices will increasingly include speech and text, potentially useful in analysis of intellectual activities. Analyzing what a person hears, reads, and sees, we should be able to measure the extent of cognitive activity devoted to a certain topic or subject by a learner. Test-based lifelog records can benefit from semantic analysis tools developed for natural language processing.  We show how semantic analysis of such text data can be achieved through the use of taxonomic subject facets and how these facets might be useful in quantifying cognitive activity devoted to various topics in a person's day. We are currently developing a method to automatically create taxonomic topic vocabularies that can be applied to this detection of intellectual activity.
**Citation**:
**Copyright**: Gregory Grefenstette, Lawrence Muchemi.
**Contact**: gregory.grefenstette@inria.fr, lawrence.githiari@inria.fr.

## 1 Introduction

The overarching goal of lifelogging is the creation, storage, analysis and eventual usage of digital records of an individual's total experience. In practice, it currently involves the passive digital capture of moments and episodes in an individual's everyday life. One objective of lifelogging is increasing self-awareness and eventual improvement of one's life. The quantity of wearable lifelogging gadgets, which track activity, physiological and environmental data, continues to grow and expand.  Alongside these quantified-self devices, body-worn video devices, e.g. helmet cameras, smartglasses, which capture image and sound, are beginning to appear. In the near future, the images they capture can be transformed into text via optical character recognition via object recognition software. Captured sound will be converted to text via automatic speech recognition.  Already, people process great quantities of text every day. These emails, social network posts, and web pages visited can also be passively captured and processed for an individual's lifelog (Hinbarji, *et al.,* 2016).

Current value-added efforts for lifelog data concentrate on analysing physical activities as opposed to cognitive activities. Analysis leading to better understanding of an individual's daily information context would benefit many research fields such as self-directed learning in online e-learning platforms. For example, a conclusion arrived at in self-directed learning environments by Guralnick (2007) establishes that an individual's information context *"influences the level of learner autonomy that is allowed in the specific context, as well as how a learner utilizes resources and strategies, and becomes motivated to learn"*.

This position paper seeks to explain how abstract text-based environments, capturable in lifelogs, might be analysed through taxonomic facets that characterise and quantify the areas of intellectual activities that a person engages in their daily life.

## 2 Present and Future of Textual Lifelog Data

In addition to quantified-self data (heart rate, steps taken, liquids and food consumed, mood, arousal, blood oxygen levels, sleep), a lifelog can increasingly contain text, sounds and images. The text to be included in a lifelog can come from four main sources: (i) digital interactions such as emails sent and received, social network posts, documents stored on a computer, web pages visited; (ii) conversion of captured, ambient speech into text via automatic speech recognition; and (iii) conversion of printed text via optical character recognition (Yi & Yingli, 2015); and (iv) the conversion of GPS coordinates into semantic descriptions of places visited (Xin, Cong & Jensen, 2010). Some numbers: A business user will receive about 75 legitimate emails per day, and send over 30 (Radicati & Levenstein, 2015). The average online user consumes over 280 posts per day amounting to 54,000 words (Bennett, 2013; Dhir & Midha, 2014). As passive conversion of speech to text continues to improve, the quantity of text to be stored on a lifelog should increase (Bellegarda & Christof, 2016). Research shows that people speak over 15,000 words per day (Mehl, 2007). A child hears 20,000 words a day (Risley & Hart, 2006), adults probably more. These observations demonstrate the enormous potential of text data that, though currently ignored, will certainly be included in future comprehensive lifelogs.

## 3   Induction of Semantic Facets in Textual Lifelog Data

In order for lifelogging to be useful as a tool for measuring cognitive activity, we will have to be able to classify a user's daily cognitive activity through natural language processing of text that they create or consume (whether it come from reading, writing, seeing, speaking or hearing). It is easy to perform word-based index textual data; it is harder to organize it into cognitive activities. And though much work has been done for capturing episodic activity (Gurrin, Smeaton, & Aiden, 2014), lifelogs do not typically capture or store cognitive activities and this will have to change (Wang, Peng, & Smeaton (2011).

Responding to this challenge of enriching diverse and massive, personal lifelog data, we have designed a private, personal search platform for capturing and classifying semantically classified cognitive data from a person's digital interactions (source (i) above, the other three sources will be treated in future versions). In their private space, a user provides credentials for their personal data sources: email and social apps, as well as quantified-self apps. This diverse data is fetched and annotated using topic vocabularies in the form of taxonomies, that the user has chosen as representing their interests. The process of inducing these taxonomies is explained in Grefenstette (2015), and in Grefenstette and Muchemi (2015 and 2016). Search facets generated from these taxonomies facilitate semantic categorising and browsing of user-generated or user-consumed data. They could also help to measure the amount of text and time devoted to certain topics, as well as the amount of topic-specific vocabulary encountered. Suppose, for example, that a student is taking a "Managerial Accounting" course. One would expect their daily activity during that period to include some reading, hearing, browsing, and speaking about topics in this field. With our taxonomy induction technique we can automatically generate a domain taxonomy such as:

        managerial_accounting>costs>variable costs>sales volume
        managerial_accounting>costs>variable costs>selling price
        managerial_accounting>costs>variable costs>transfer price
        managerial_accounting>financing
        managerial_accounting>financing>activity-based costs
        managerial_accounting>financing>business decision
        managerial_accounting>financing>financial accounting
        managerial_accounting>financing>financial accounting>bookkeeping

Figure 1. An Example of the Induced taxonomy for the topic "Managerial Accounting"

This taxonomy includes a rich vocabulary related to the domain (*…, financial ratios, financial report, financial reporting, financial reports, financials, financial statement, financial statement analysis, financial statements, find results, fixed cost, fixed costs, gaap, garrison, general accepted accounting principles, generally accepted accounting principles, graduate certificate, historical cost, historical costs, …* ) that can be used to annotate lifelog entries as belonging to this topic, once the topic taxonomy is activated by the user. In our ongoing work we have tested our induced taxonomies to successfully distinguish topics in text sources such Reddit comments. We have created hundred of taxonomies for personal activities such as hobbies and illnesses. The results will be publicly available as soon as our experiments are complete, but in the meantime, it seems that it is feasible to easily create a large number of targeted vocabularies, and that these vocabularies can be used to classify daily activity into domains which can be used to measure the actual cognitive activity of future lifeloggers, just as quantified self tools can be used to measure physical activities today.

## 4   Conclusion

The partitioning of text-based lifelog data using domain taxonomies can facilitate analyzing of lifelog and classifying activity retrospectively. Though some attempts at developing systems that allow manual classification of lifelogged activity have been proposed (MyLifeBits (Gemmell, Bell & Lueder, (2004), LifeLog (Kiyoharu, *et al.* 2004), Stuff I'veSeen (Dumais, *et al.* 2003), PERSONE (Kim, *et al.* 2006) Personal Data Prototype (Teraoka, 2012)), we feel that activity annotation must be an automatic and passive process. Loggerman (Hinbarji, *et al.,* 2016) is a recent system that allows automatic logging of a person's typing and app use. This is a good start, but we believe each piece of information that a user generates or consumes must also be semantically classified and annotated. Annotating a person's cognitive activity will allow the user, and anyone that the user shares their data with, to judge whether time spent learning is sufficient, and pooling the results of users will provide an additional dimension for improving directed and self-directed learning.

## 5  References

Bellegarda, J., & Christof, M. (2016). "*State of the art in statistical methods for language and speech processing*." Computer Speech & Language 35: 163-184.

Bennett, S. (2013). *Social Media Overload – How Much Information Do We Process Each Day?* . (A. B. Network, Producteur) Consulté le January 2016, sur Social Times: http://www.adweek.com/socialtimes/social-media-overload/488800

Dhir, M., & Midha, V., (2014). *"Overload, Privacy Settings, and Discontinuation: A Preliminary Study of FaceBook Users"*. SIGHCI 2014 Proceedings. Paper 12 ; and Bennett, S. (2013, July 31). Social Media Overload – How Much Information Do We Process Each Day? . (A. B. Network, Producteur) Consulted  January 2016, on Social Times: http://www.adweek.com/socialtimes/social-media-overload/488800

Dumais, S., Cutrell, E., Cadiz, J., Jancke, G., Sarin, R., & Robbins, D. (2003). *"Stuff I've Seen: A System for Personal Information Retrieval and Re-Use. SIGIR '03":* Proceedings of the 26th annual international ACM SIGIR conference on Research and Dev. in information retrieval. 72–79. New York, NY, USA.

Gemmell, J., Bell, G., & Lueder, R. (2004). "*MyLifeBits: a personal database for everything*." Communications of the ACM (CACM) , 49 (1), 88-95.

Grefenstette, G. (2015). "*Simple Hypernym Extraction Methods.*" HAL-INRIASAC. Palaiseau, France

Grefenstette, G., & Muchemi, L. (2015). "*Extracting Hierarchical Topic Models from the Web for Improving Digital Archive Access. Expert Workshop on Topic Models and Corpus Analysis*-DARIAH Text & Data Analytics Working Group (TDAWG). Dublin-Ireland.

Grefenstette, G., & Muchemi, L. (2016). Automatic Taxonomy Construction using a Domain Crawler and Sentence Level Phrase Co-occurrence Heuristics.

Guralnick, D. (2007). The Importance of the Learner's Environmental Context in the Design of M-Learning Products. 2nd International Conference on "Interactive Mobile and Computer aided Learning". Amman, Jordan.

Gurrin, C., Smeaton, A., &  Aiden, D., (2014). LifeLogging: Personal Big Data, Foundations and Trends in Information Retrieval. Vol. 8, No. 1, 1–107

Hinbarji, Z., Albatal, R., O'Connor, N. and Gurrin, C., 2016, January. LoggerMan, a Comprehensive Logging and Visualization Tool to Capture Computer Usage. In MultiMedia Modeling (pp. 342-347). Springer International Publishing.

Kim I., J., Ahn S.,C., Ko H., Kim H.,G. (2006) PERSONE: Personalized Experience Recoding and Searching On Networked Environment. Proceedings of the 3rd ACM Workshop on Continuous Archival & Retrieval of Personal Experiences: 27 Oct 2006; Santa Barbara, USA. ACM pp 49–54

Kiyoharu, A., Hori, T., Kawasaki, S., & Ishikawa, T. (2004). Capture and Efficient Retrieval of Life log. Pervasive 2004 Workshop on Memory and Sharing Experiences, (pp. 15-20).

Mehl, M., Vazire, S., Ramírez-Esparza, N., Slatcher, R. B., & Pennebaker, J. (2007). Are Women Really More Talkative Than Men? Science Magazine , 317, 82.

Radicati and Levenstein.( 2015).  Email statistics report, 2015-2019. Technical report, 2015.

Risley, T. R & Hart, B. (2006). Promoting early language development. In N. F. Watt, et al (Eds.), The crisis in youth mental health: Critical issues and effective programs, Volume 4, Early intervention programs and policies (pp. 83-88). Westport, CT

Teraoka Teruhiko (2012). "Organization and exploration of heterogeneous personal data collected in daily life." Human-Centric Computing and Information Sciences 2:1 1-15.

Wang, Peng, and Alan F. Smeaton (2011). "Aggregating Semantic Concepts for Event Representation in Lifelogging." Proceedings of the International Workshop on Semantic Web Information Management. ACM

Xin, C., Cong, G., & Jensen. C. (2010). "Mining significant semantic locations from GPS data." Proceedings of the VLDB Endowment 3.1-2: 1009-1020.

Yi, C., & Yingli, T. (2015). "Assistive Text Reading from Natural Scene for Blind Persons." Mobile Cloud Visual Media Computing. Springer International Publishing, 2015. 219-241.